\documentclass[preprint,12pt,authoryear]{elsarticle}

\usepackage{amssymb}

\usepackage{makecell}

\usepackage{algorithm}
\usepackage{algorithmic}

\usepackage{multirow}

\usepackage{pifont}

\usepackage{graphicx}
\usepackage{amsmath}
\usepackage{amssymb}
\usepackage{booktabs}

\usepackage{color}
\usepackage{setspace}
\usepackage{hyperref}
\usepackage[hyphenbreaks]{breakurl}

\usepackage{geometry}
\usepackage[capitalize]{cleveref}
\crefname{section}{Sec.}{Secs.}
\Crefname{section}{Section}{Sections}
\Crefname{table}{Table}{Tables}
\crefname{table}{Tab.}{Tabs.}

\begin{document}

\begin{frontmatter}



\title{Multi-View Black-Box Physical Attacks on Infrared Pedestrian Detectors Using Adversarial Infrared Grid}

\author[label1]{Kalibinuer Tiliwalidi}
\ead{202011081727@std.uestc.edu.cn}

\author[label2]{Chengyin Hu\corref{cor1}}
\ead{cyhuuestc@gmail.com}



\author[label1]{Weiwen Shi}
\ead{weiwen_shi@foxmail.com}






\affiliation[label1]{organization={University of Electronic Science and Technology of China},
            addressline={No. 2006, Xiyuan Avenue, Gaoxin District},
            city={Chengdu},
            postcode={611731},
            state={Sichuan},
            country={China}}

\affiliation[label2]{organization={China University of Petroleum-Beijing at Karamay},
            addressline={NO. 355, Anding Road},
            city={Karamay},
            postcode={834000},
            state={Xinjiang},
            country={China}}

\cortext[cor1]{Corresponding author}
%

\begin{abstract}

While extensive research exists on physical adversarial attacks within the visible spectrum, studies on such techniques in the infrared spectrum are limited. Infrared object detectors are vital in modern technological applications but are susceptible to adversarial attacks, posing significant security threats. Previous studies using physical perturbations like light bulb arrays and aerogels for white-box attacks, or hot and cold patches for black-box attacks, have proven impractical or limited in multi-view support. To address these issues, we propose the Adversarial Infrared Grid (AdvGrid), which models perturbations in a grid format and uses a genetic algorithm for black-box optimization. These perturbations are cyclically applied to various parts of a pedestrian's clothing to facilitate multi-view black-box physical attacks on infrared pedestrian detectors. Extensive experiments validate AdvGrid's effectiveness, stealthiness, and robustness. The method achieves attack success rates of 80.00\% in digital environments and 91.86\% in physical environments, outperforming baseline methods. Additionally, the average attack success rate exceeds 50\% against mainstream detectors, demonstrating AdvGrid's robustness. Our analyses include ablation studies, transfer attacks, and adversarial defenses, confirming the method's superiority.

\end{abstract}



\begin{keyword}

Infrared object detectors, Adversarial infrared grid, Genetic algorithm, Multi-view black-box physical attacks

\end{keyword}

\end{frontmatter}



\section{Introduction}

Currently, deep neural networks (DNNs) have achieved remarkable success across various fields and have become a core technology in artificial intelligence. They are widely applied with significant effectiveness in areas such as image recognition \cite{ref1}, speech recognition \cite{ref2}, natural language processing \cite{ref3}, and object detection \cite{Yolov3}. For instance, in image recognition, DNNs can accurately classify and identify various images; in speech recognition, they can convert speech into text, achieving high-precision speech-to-text conversion; in natural language processing, DNNs can understand and generate natural language, enabling functionalities such as automatic translation and question-answering systems; in object detection, DNNs can accurately identify and locate target objects in images and videos.

Regarding object detection, many research efforts involve training detectors by collecting a large number of samples in visible light environments \cite{Mask, faster}. These detectors, carefully designed and trained, demonstrate strong robustness and high accuracy under standard lighting conditions. However, their performance is significantly affected in low-light or dim conditions. To overcome this limitation, some researchers  \cite{ref40} have recently shifted their focus to the infrared domain. By collecting infrared samples, researchers can train robust detectors that perform well under various lighting conditions. These infrared detectors not only function normally during the day but also maintain high-efficiency target detection capabilities at night or in low-light environments. The success of this approach not only broadens the application scope of object detection technology but also provides new solutions for addressing illumination variations in practical applications.

As DNN-based tasks advance, their security performance has also garnered scholarly attention. Infrared target detectors, combining infrared imaging technology and deep neural networks, are shown to be vulnerable to adversarial attacks \cite{Lichao1,ref6}. In adversarial attacks, digital adversarial attacks \cite{Lichao2} involve adding imperceptible perturbations directly to digital images. In contrast, physical adversarial attacks \cite{ref9, ref10} require adding noticeable perturbations to the target object, then using sensors to capture samples and input them into the target model to execute the attack. Most current physical adversarial attacks \cite{FCA,DAS} on target detectors are deployed in the visible light modality, with only a few studies \cite{Bulbattack,AdvCloth} exploring physical attacks in the infrared modality, making the research on infrared physical adversarial attack techniques particularly urgent.

\begin{figure*}
\centering
\includegraphics[width=1\linewidth]{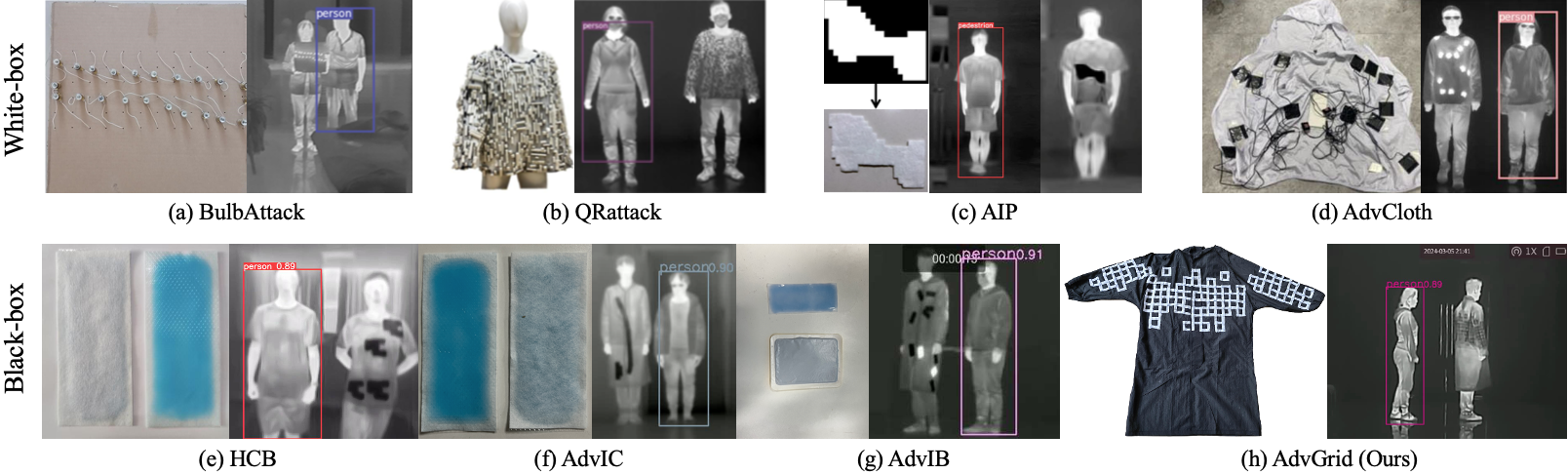} 
\caption{Existing infrared physical adversarial attack methods vs. proposed method.}
\vspace{-0.3cm}
\label{figure1}
\end{figure*}

Existing infrared physical adversarial attack techniques, as illustrated in Figure \ref{figure1}, encompass various approaches utilizing different physical perturbations to execute infrared attacks. These include using light bulb arrays Figure \ref{figure1}(a) \cite{Bulbattack}, QR code clothing Figure \ref{figure1}(b) \cite{QRattack}, aerogels Figure \ref{figure1}(c) \cite{AIP}, and electric heating films Figure \ref{figure1}(d) \cite{AdvCloth}. These methods primarily target white-box physical attacks on infrared pedestrian detectors, operating under the assumption that the attacker has access to the target model's architecture and parameters. While this assumption holds in theoretical research, it is impractical in real-world scenarios, where attackers typically cannot obtain such detailed internal information. For black-box physical attacks on infrared pedestrian detectors \cite{AdvIB, AdvIC, HCB}, researchers often employ hot and cold pastes as physical perturbations Figures \ref{figure1}(e), (f), (g). These materials are readily available and can be discreetly affixed inside a pedestrian's clothing, making them difficult for human observers to detect. Without infrared sensors, defenders struggle to identify the presence of these physical perturbations. However, these existing black-box attack methods have significant limitations, particularly in achieving multi-view adversarial effects. This means that the attack's effectiveness diminishes or fails entirely when the target is observed or detected from different angles, thus limiting the practical applicability and effectiveness of the attack.

To address the limitations of existing infrared physical adversarial attack methods, this study proposes the 
AdvGrid for multi-view black-box adversarial attacks on infrared pedestrian detectors. This method utilizes cold patches that can be affixed inside clothing as physical perturbations to enhance stealth. By simulating the grid model and optimizing its physical parameters using genetic algorithms \cite{GA}, the most adversarial infrared grid parameters are obtained. The Expectation Over Transformation 
framework \cite{EOT} is then employed for robust enhancement, further improving the adversarial effect. Additionally, Thin Plate Spline 
\cite{TPS} is used for fold simulation to enhance the stability of physical perturbations. Finally, in the physical world, the optimized perturbations are cyclically attached to the inside of pedestrian clothing, achieving multi-view black-box attacks on infrared pedestrian detectors.

\begin{table*} [h]
	\centering
 
    \setlength{\belowcaptionskip}{5pt}
    \caption{Performance comparison between existing infrared attack techniques.}
    \label{tab1}
	\begin{tabular}{ccccc}

    \hline
    ~&Perturbation&Stealthiness&Scenario&Multi-view\\
    \hline
    BulbAttack&Bulbs&\ding{55}&White-box&\ding{55}\\

    QRattack&Aerogels&\ding{55}&White-box&\ding{51}\\
    
    AIP&Aerogels&\ding{55}&White-box&\ding{55}\\

    AdvCloth&Electric heating films&\ding{51}&White-box&\ding{55}\\
    
    HCB&Hot and cold pastes&\ding{51}&Black-box&\ding{55}\\

    AdvIC&Cold pastes&\ding{51}&Black-box&\ding{55}\\

    AdvIB&Hot and cold pastes&\ding{51}&Black-box&\ding{55}\\

    AdvGrid (Ours)&Cold pastes& \ding{51} &Black-box&\ding{51}\\

    \hline

    \end{tabular}
\vspace{0.3cm}
\end{table*}

Table \ref{tab1} compares the proposed method with existing infrared physical attack methods. It is evident that besides AdvGrid, no other method simultaneously offers the advantages of black-box adversarial attacks, stealth, and multi-view capability. Additionally, the cost of deploying physical adversarial attacks using our method is inexpensive, with a total cost of less than \$10, making it easier to implement. The contributions of our method are summarized as follows:

\begin{itemize}

\item \textbf{AdvGrid: A New Physical Attack on Infrared Pedestrian Detectors:} 
AdvGrid introduces a cost-effective, highly stealthy, black-box attack using cold patches. It effectively targets models without prior knowledge and maintains efficient attack performance from multiple angles, posing a real-world security threat at a cost of less than \$10.

\item \textbf{Extensive Experimental Validation:} Comprehensive experiments confirm AdvGrid's superior effectiveness, stealth, and robustness across various test environments. It outperforms baseline methods, successfully performing undetected adversarial attacks on infrared pedestrian detectors  without being detected by the naked eye, while maintaining stable attack performance at various distances and angles.

\item \textbf{In-depth Method Analysis:} 
Ablation studies and transfer attack experiments validate the contribution of each component and the generality of AdvGrid across different detectors. Adversarial defense experiments offer effective countermeasures, reinforcing AdvGrid's leading position in physical attacks on infrared pedestrian detectors.


\end{itemize}

\section{Related works}

Given that the proposed method targets object detectors, this section will focus on physical attacks against object detectors, detailing techniques in both visible and infrared modalities.

\subsection{Physical attacks in the visible light field}

Pedestrian object detection technology plays a key role in enhancing quality of life, urban management, and intelligent development. Its primary task is to accurately detect and locate pedestrians in images or videos. 

Xu et al. proposed AdvTshirt \cite{AdvTshirt}, which uses flexible transformation techniques to simulate clothing deformation for robust optimization, ensuring the robustness of physical perturbations on non-rigid objects. Employing minimax optimization, this work extends to integrated attacks against multiple object detectors simultaneously. Experimental results validate the robustness of AdvTshirt across different pedestrian object detectors.

Wu et al. introduced AdvCloak \cite{AdvCloak}, which trains perturbation patterns using standard datasets to suppress widely used object detectors, thereby enhancing adversarial attack transferability. Experimental results confirm its effectiveness in both white-box and black-box settings, as well as its transferability across datasets, object categories, and detectors.

Tan et al. proposed LAP \cite{LAP}, aimed at deceiving both human vision and object detectors. It employs a dual-stage training strategy to generate perturbation patterns that appear plausible to humans while fooling object detectors. Experimental results demonstrate that LAP achieves significant adversarial effects and maintains a reasonable visual appearance.

Hu et al. introduced NP \cite{NP}, which focuses on the visual plausibility of perturbations to enhance attack stealth. This work utilizes a pretrained generative adversarial network to learn real-world images, sampling optimal images to produce natural adversarial patches under high adversarial conditions. Experimental results indicate superior adversarial effects, with independent subjective surveys showing that NP-generated adversarial samples appear more natural than baseline methods.

To address multi-view adversarial attacks, Hu et al. proposed AdvTT \cite{AdvTT}, based on a scalable generative framework with circular cropping, creating adversarial textures with repetitive structures. These textures are deployed on all sides of clothing to achieve multi-view adversarial attacks. Experimental results demonstrate that clothing items like t-shirts and skirts with these textures effectively execute multi-view adversarial attacks against pedestrian object detectors.

Vehicle object detectors are crucial in computer vision, tasked with accurately detecting and locating vehicles in images or videos. Applications include traffic monitoring, autonomous driving, and security. 

Zhang et al. introduced CAMOU \cite{CAMOU}, which trains a neural approximation function to simulate camouflaged vehicles, searching for optimal camouflage to minimize approximate detection scores. Experimental results show that this camouflage deceives advanced vehicle detectors and generalizes across different vehicles, environments, and detectors.

Huang et al. proposed UPC \cite{UPC}, which introduces a set of transformations to simulate deformable properties, making the generated camouflages effective for non-rigid or non-planar objects. Experimental results indicate advantages in both virtual and physical environments.

Wang et al. introduced DAS \cite{DAS}, which simultaneously suppresses model and human attention, generating highly transferable and visually natural physical adversarial camouflages. Extensive experiments in digital and physical worlds against state-of-the-art models demonstrate superior performance over existing methods.

Wang et al. proposed FCA \cite{FCA}, rendering non-planar camouflage textures over the entire vehicle surface, and introducing a transformation function to convert optimized vehicles into more realistic scenes. Experimental results show that full-coverage camouflage attacks outperform existing methods across various test cases and generalize to different environments, vehicles, and detectors.

Duan et al. introduced CAC \cite{CAC}, which transforms camouflage based on object posture in 3D space, fixes the top $n$ proposals of the region proposal network, and attacks all classifications within fixed dense proposals to output errors. Experiments demonstrate that CAC outperforms existing attack algorithms, showing superior performance in both virtual and real-world scenarios.

\subsection{Physical attacks in the infrared field}

Zhu et al. pioneered physical adversarial attacks in the infrared modality with BulbAttack \cite{Bulbattack}, using a set of light bulbs as heat sources to create white perturbations under infrared imaging. This optimization deceives well-trained infrared detectors, as demonstrated by an individual holding the optimized bulb board to evade detection.

Subsequently, Zhu et al. proposed QRattack \cite{QRattack}, which uses aerogel as insulation to create black perturbations in thermal imaging. Optimized adversarial QR codes were manually simulated, and individuals wearing these QR code vests could avoid infrared detection.

Wei et al. advanced the field with HCB \cite{HCB}, utilizing heat and ice patches as perturbations optimized with evolutionary algorithms for black-box attacks on infrared detectors. These patches, hidden inside clothing, made it difficult for humans to detect.

Wei et al. also introduced AIP \cite{AIP}, using aerogel patches optimized through a polynomial regularization method. Experimentally validated, these patches adhered to the fronts of clothing and effectively evaded infrared detectors.

Zhu et al. developed AdvCloth \cite{AdvCloth}, employing flexible carbon fiber heaters as perturbations, constrained by L\_dist loss functions for deployment feasibility in the real world. Experimental results showed significant improvements in both digital and physical adversarial effects.

Hu et al. proposed AdvIC \cite{AdvIC}, using Bezier curves to model infrared perturbations, optimized with particle swarm algorithms. Effective real-world adversarial attacks were demonstrated using only two Bezier curves.

Hu et al. further developed AdvIB \cite{AdvIB}, modeling perturbations as discrete patches with hot and cold stickers, optimized with differential evolution algorithms. Extensive experiments validated the multi-distance and transfer attack effectiveness in the physical world.

\section{Methodology}

\subsection{Problem definition}

\textbf{Object detector:} Consider an infrared pedestrian imaging dataset $DS$, where $X$ denotes the clean infrared images and $Y$ represents the corresponding correct labels for pedestrians. Let $f$ denote the object detector. For each input image $X \in DS$ in the dataset, the pretrained model $f : X \rightarrow Y$ of the object detector analyzes the image and predicts a label $y$ that aligns with the correct label $Y$. The predicted label $y$ includes the following key components: the location information of the bounding box (${y}_{pos}$), the confidence or probability of the target (${y}_{obj}$), and the category information of the target (${y}_{cls}$). Utilizing this information, the object detector can accurately identify and locate pedestrians, ensuring stable and reliable detection performance across various environments:

\begin{equation}
    \label{for1}
    y:=[{y}_{pos},{y}_{obj},{y}_{cls}]=f(X)
\end{equation}

\textbf{EOT Framework:} The Expectation Over Transformation (EOT) framework \cite{EOT} presents a systematic approach to generate resilient adversarial samples, with the aim of bolstering the efficacy of adversarial attacks across diverse environments and conditions. Originally introduced by Athalye et al. in 2018, this framework has found extensive application in the realm of adversarial attack research within physical contexts. Leveraging a series of stochastic transformations applied to input samples, the EOT framework ensures the resilience of generated adversarial samples across varying conditions. Central to its methodology is the optimization of adversarial samples to effectively subvert target models amidst multiple transformational scenarios, encompassing translations, rotations, scalings, and alterations in lighting conditions. While conventional adversarial attack methods typically operate under the assumption of static and unchanging input data, real-world scenarios often entail dynamic transformations. EOT, by accounting for these transformations, engenders adversarial samples of heightened robustness.

\textbf{TPS Framework:} The Thin Plate Spline (TPS) framework \cite{TPS} constitutes a widely employed deformation model prevalent in image processing, computer vision, and machine learning domains. Originally formulated by Fred L. Bookstein in 1989, it primarily addresses non-rigid deformations of images and shapes. The TPS model employs a smooth and differentiable function to effectuate the transformation from a source image or shape to a target counterpart. Its procedural steps entail the selection of control points, adoption of the TPS deformation model, solution of linear equations to ascertain deformation function parameters, pixel-wise mapping from the source to the target image, and eventual application of the deformation function to yield the deformed image.

\textbf{Genetic Algorithm:} The Genetic Algorithm (GA) \cite{GA} constitutes an optimization paradigm rooted in principles of natural selection and genetics, initially conceptualized by John Holland in the 1970s. Emulating biological evolution, GA orchestrates operations such as selection, crossover, and mutation to iteratively refine solutions. This algorithmic approach finds widespread utility in tackling intricate optimization conundrums, particularly those resistant to conventional methodologies. Noteworthy attributes encompass potent global search capabilities, adaptability, and versatility in addressing diverse optimization problem typologies.

\subsection{Adversarial attack framework}

\begin{figure*}[h]
\centering
\includegraphics[width=1\linewidth]{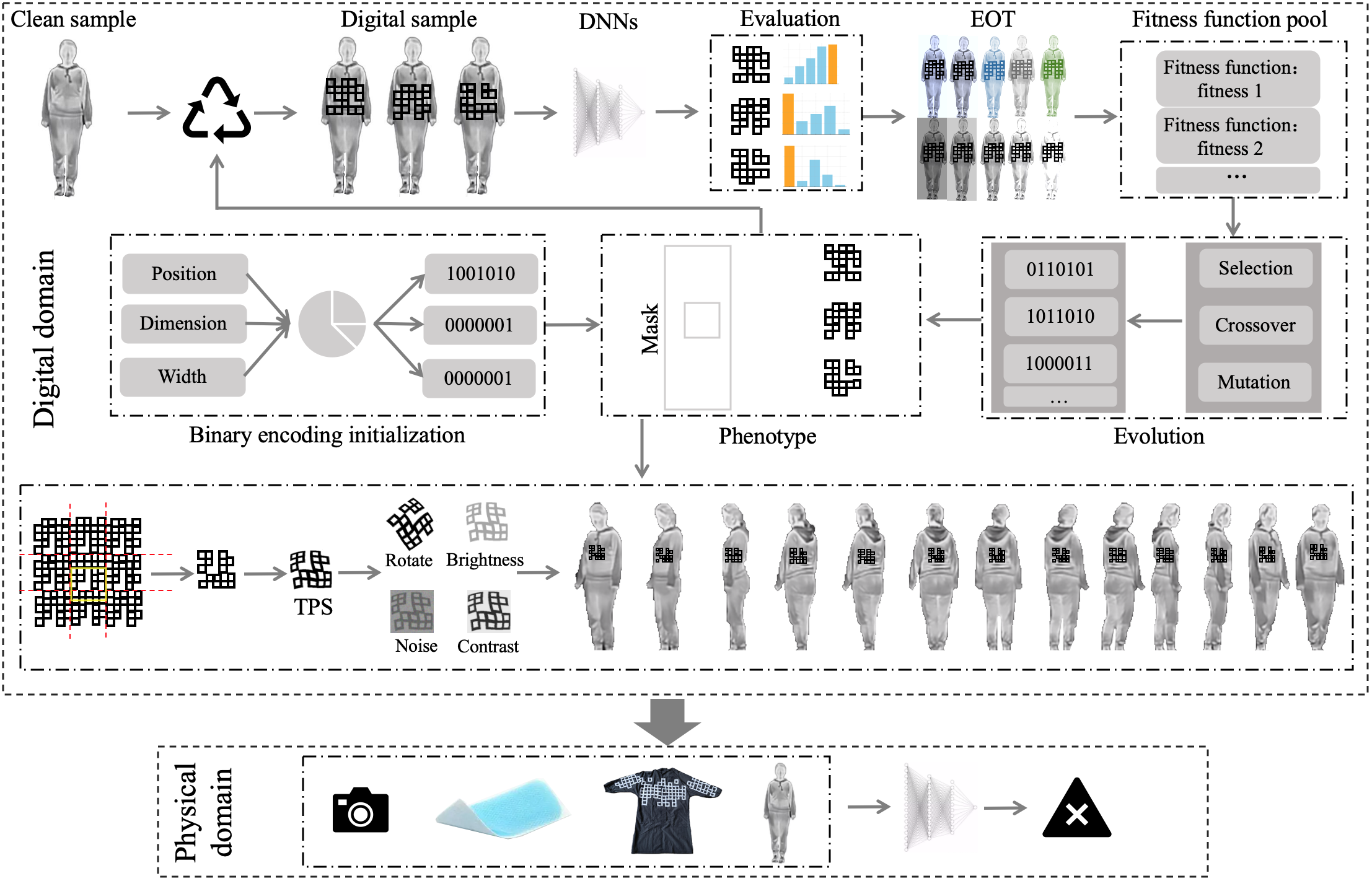} 
\caption{AdvGrid's adversarial attack framework. The AdvGrid attack encompasses a multifaceted process primarily conducted within the digital domain, involving stages such as simulation, optimization, and robustness enhancement. Subsequently, in the physical domain, the attack strategy involves fabricating infrared suits augmented with cold patches to serve as disruptive elements. These suits are then utilized to execute the attack, leveraging the physical manifestation of perturbations to undermine targeted systems.}
\vspace{-0.3cm}
\label{figure2}
\end{figure*}

Figure \ref{figure2} illustrates the adversarial attack framework of this method, comprising seven key components:
\textbf{Simulation and Modeling:} This initial phase involves modeling the infrared grid to ascertain its physical parameters, encompassing factors like location, dimensions, and width. These parameters serve as the foundation for subsequent optimization steps.
\textbf{Binary Encoding and Population Initialization:} The physical parameters of the infrared grid are encoded into binary form, and an initial population is initialized. This step ensures diversity and a rational distribution within the population.
\textbf{Conversion to Digital Perturbations:} Binary encoded parameters are transformed into digital perturbations, which are then combined with clean samples to generate digital representations simulating real-world physical perturbations.
\textbf{Evaluation and EOT Robustness Enhancement:} The digital samples are input into DNNs for evaluation. Subsequently, EOT robustness enhancement techniques are applied to augment the stability and efficacy of the adversarial perturbations.
\textbf{Genetic Algorithm Optimization:} A suitable fitness function is selected, guiding the selection of individuals within the population. Crossover and mutation operations are performed iteratively on the population using genetic algorithms, gradually refining parameters towards optimal solutions.
\textbf{Perturbation Splicing and TPS Robustness Enhancement:} The most adversarial perturbations are selected, followed by random cropping of perturbation blocks. TPS robustness enhancement techniques are applied to enhance stability and resistance to deformation. Enhanced perturbations are then integrated into pedestrian images from various viewpoints, ensuring effective attacks from multiple perspectives.
\textbf{Physical Deployment:} In the physical realm, optimized perturbation patterns are embedded within pedestrian clothing. This facilitates multi-view adversarial effects, rendering real pedestrians challenging to detect by infrared pedestrian detectors in practical scenarios, thereby achieving covert attack objectives.

\subsection{Generate adversarial sample}

In this work, infrared perturbations are simulated and modeled using a grid approach for multi-view optimization and deployment. As shown in Figure \ref{figure3}, each individual infrared grid block consists of four physical parameters: location, color, width, and dimension. The detailed definitions of each parameter are as follows:

\begin{figure}[h]
\centering
\includegraphics[width=1\columnwidth]{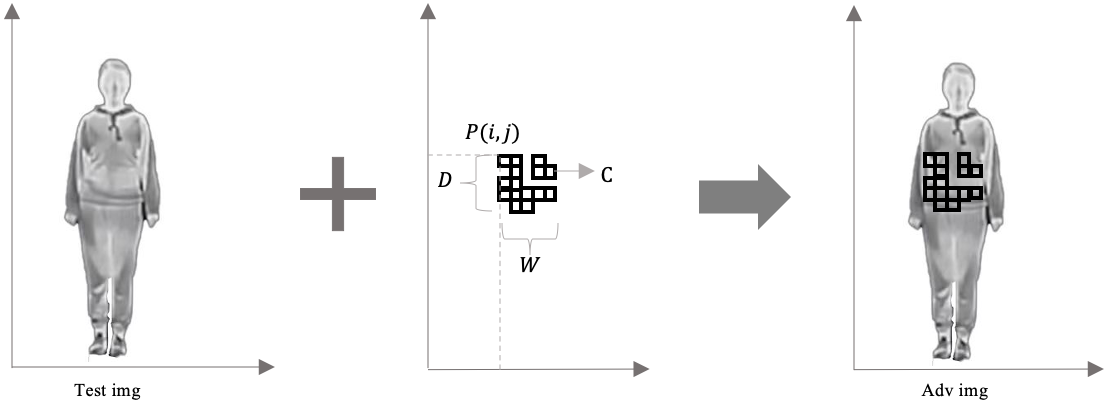} 
\caption{Infrared grid simulation modeling diagram}
\vspace{-0.3cm}
\label{figure3}
\end{figure}

\textbf{Location ($P(i,j)$):} In this study, $P(i,j)$ represents the top-left coordinate of each independent grid, determining the specific location of the grid on the pedestrian's body. Each independent grid within the grid block is continuously simulated, meaning the grids are connected. Disconnected appearances in Figure \ref{figure3} are due to transparent grid settings (i.e., no perturbation at that position). Thus, the position parameters are formalized as $P = {({i}_{11}, {j}_{11}), ({i}_{12}, {j}_{12}), ..., ({i}_{DD}, {j}_{DD})}$, where $(i11, j11)$ represents the position of each independent subgrid in the simulated infrared grid block.

\textbf{Color ($C$):} Parameter $C$ represents the color of the infrared grid block. Since objects under infrared cameras typically appear in black, white, and gray, the color value of each subgrid perturbation in this study is fixed to either no color or black ($C(0,0,0)$). When the subgrid color is set to no color, the perturbation at that position becomes discontinuous, increasing the diversity of perturbations and enhancing the likelihood of successful adversarial attacks.

\textbf{Width ($W$):} Parameter $W$ represents the width of the infrared grid. If the grid width is too large, it will exceed the pedestrian's boundary; if the width is too small, the subgrids will be too dense, which is not conducive to physical deployment. After repeated testing, setting the width to one-fifth of the height of the pedestrian's bounding box is found to be a reasonable configuration. This setting is advantageous because in digital images where only the left or right half of the pedestrian's body is captured, setting the width to two-thirds of the bounding box width results in a smaller perturbation area, and the generated perturbations do not achieve the expected effect.

\textbf{Dimension ($D$):} Parameter $D$ represents the dimension of the infrared grid. After fixing the grid width, if the dimension is set too large, the grids will be too dense, making it difficult to deploy in the physical world. Conversely, if the dimension is set too small, it reduces the diversity of perturbations and weakens the effectiveness of digital adversarial attacks. A small dimension also causes large wrinkles during multi-view deployment, weakening the physical attack effect. In the experiment, the dimension is set to $D = 8$ to ensure that the perturbations are diverse and can be effectively deployed in the physical world.

We use $\theta(P, C, W, D)$ (abbreviated as $\theta$) to represent an infrared grid. To ensure that the perturbations appear in the pedestrian area, a mask $\mathcal{M}$ is used to limit the position area of the perturbations. Therefore, the process of generating adversarial samples can be represented as follows:

\begin{equation}
    \label{for2}
    {X}_{adv}=S(X,\theta,\mathcal{M})
\end{equation}

In this context, ${X}_{adv}$ represents the adversarial sample, and $S$ is a linear fusion function that combines the clean sample with the simulated infrared grid to create the adversarial sample.

After generating the adversarial samples, it is important to address the challenges associated with transferring these samples from a static and fixed digital environment to the physical domain. This transfer can introduce issues such as position errors, lighting changes, and noise impacts. To mitigate these issues, the EOT framework \cite{EOT} is employed to simulate various transformation conditions, thereby generating more robust and practical adversarial samples. The EOT framework uses a distribution of transformations $\mathcal{T}$ to model domain transfer, where each instance represents a combination of random image transformations, including perspective transformation, brightness adjustment, downsampling, and more. Consequently, the adversarial samples enhanced by the EOT framework can be expressed as follows:

\begin{equation}
    \label{for3}
    {X}_{adv} = {\mathbb{E}}_{t \sim \mathcal{T}}t({X}_{adv})
\end{equation}

Following robustness enhancement with the EOT framework, it is necessary to consider the deployment of perturbations inside pedestrian clothing. Given that clothing is a non-rigid object prone to deformation, the TPS architecture \cite{TPS} is further employed to map the perturbations from the digital domain to the physical domain. This ensures that the enhanced adversarial samples can better adapt to changes in the real environment. Specifically, the enhanced adversarial samples can be represented as:

\begin{equation}
    \label{for4}
    {X}_{adv} = TPS({\mathbb{E}}_{t \sim \mathcal{T}}t({X}_{adv}))
\end{equation}

\subsection{Infrared grid adversarial attack}

The proposed method aims to optimize the physical parameters of the infrared grid using a genetic algorithm to generate the most adversarial infrared grid, which is then fused with clean samples to obtain the most adversarial digital samples. This method targets more realistic scenarios for black-box attacks, where the attacker cannot access the internal information of the model such as its architecture and parameters, and can only obtain the model's output, such as the detected object's category and confidence. Therefore, the method takes the confidence of the target object as the adversarial loss and formalizes the objective function to minimize the confidence of the infrared detector for the target object:

\begin{equation}
    \label{for5}
    \mathop{\arg\min}_{\theta}TPS({\mathbb{E}}_{t \sim \mathcal{T}}({y}_{obj} \leftarrow f(t({X}_{adv}))))
\end{equation}

This method implements the global optimization shown in Equation \ref{for5} using a genetic algorithm to obtain the final optimal solution, which are the physical parameters of the most adversarial infrared grid. These physical perturbations are then deployed in the real world to deceive infrared target detectors. The process of optimizing the infrared grid using a genetic algorithm \cite{GA} is introduced below.

\textbf{Random Initialization:} During the optimization process of AdvGrid, a randomly initialized population of candidate solutions ($POP$) is first generated. Each solution in the population corresponds to a set of physical parameters $\theta$ that define the infrared grid used to generate adversarial samples. Random initialization helps ensure the diversity of the search space, increasing the likelihood of finding the optimal solution. This process is represented as:

\begin{equation}
    \label{for6}
    POP=[{\theta}_{1},{\theta}_{2},...,{\theta}_{G}]
\end{equation}
where $G$ represents the population size and ${\theta}_{g}$ represents a candidate solution in the population ($g=1,2,...,G$).

\textbf{Fitness Calculation:} After initializing the population, each individual in the population generates adversarial samples. The confidence score of the model's output for the correct label on these adversarial samples is used as the fitness value for each individual. This way, the effectiveness of each individual in attacking the target model can be evaluated—the higher the fitness value, the more successful the adversarial sample is:

\begin{equation}
    \label{for7}
    {fit}_{g}=1-({y}_{obj}^{g} \leftarrow f(S(X,{\theta}_{g},\mathcal{M}))) \quad \quad g=1,2,...,G
\end{equation}
where, ${fit}_{g}$ represents the fitness value of the $g-th$ individual.

\textbf{Selection Operation:} Based on the confidence values of each individual in the population, a selection operation is performed. This work adopts a novel selection operation where individuals with confidence values greater than 80\% (i.e., fitness less than 20\%) are eliminated first, as these individuals perform the worst in the attack. New individuals are then randomly generated to replenish the population, maintaining diversity and population size. The selection operation is denoted by $SL$, and the new population after selection can be represented as:

\begin{equation}
    \label{for8}
    {POP}_{s+1} \leftarrow  SL({POP}_{s}, {fit}_{g}) \quad \quad s=1,2,...,S \quad g=1,2,...,G
\end{equation}
where $S$ represents the total number of generations, and ${POP}_{s}$ represents the population in the $s-th$ generation.

\textbf{Crossover Operation:} First, genes in the population are randomly paired. Then, for each pair of genes, a random gene position is selected, and with a probability ${P}_{c}$, the selected gene and the genes following it are exchanged, achieving the crossover operation. The crossover operation is denoted by $CS$, and the new gene pairs after crossover can be represented as:

\begin{equation}
    \label{for9}
    {POP}_{s+1} \leftarrow  CS({POP}_{s}, {P}_{c}) \quad \quad s=1,2,...,S
\end{equation}

\textbf{Mutation Operation:} After crossover, the population undergoes mutation. For each individual in the population, a gene is randomly selected, and with a probability ${P}_{m}$, this gene is flipped, i.e., 0 is mutated to 1 and 1 is mutated to 0. The mutation operation is denoted by $MT$, and the mutated population can be represented as:

\begin{equation}
    \label{for10}
    {POP}_{s+1} \leftarrow  MT({POP}_{s}, {P}_{m}) \quad \quad s=1,2,...,S
\end{equation}

AdvGrid, through $S$ iterations of the genetic algorithm, obtains the physical parameters of the most adversarial infrared grid to generate adversarial samples that can deceive the infrared target detector. These samples are then robustly enhanced using the EOT framework and TPS. Algorithm \ref{algorithm1} shows the pseudocode of the proposed AdvGrid, with clean samples $X$, target detector $f$, population size $G$, number of iterations $S$, crossover rate ${P}_{c}$, and mutation rate ${P}_{m}$ as inputs. The algorithm randomly initializes the population and performs fitness calculation, crossover, and mutation operations for each individual, repeating this loop continuously. The algorithm ultimately outputs the physical parameters ${\theta}^{\star}$ for subsequent physical attack experiments.

\begin{algorithm}[h]
	\renewcommand{\algorithmicrequire}{\textbf{Input:}}
	\renewcommand{\algorithmicensure}{\textbf{Output:}}
	\caption{Pseudocode of AdvGrid}
	\label{algorithm1}
	\begin{algorithmic}[1]
	
		\REQUIRE Clean sample $X$, Detector $f$, Population size $G$, Iterations $S$, Hyperparameters of GA: ${P}_{c}$, ${P}_{m}$;
		\ENSURE Physical parameters ${\theta}^{\star}$;

		\STATE \textbf{Initialization} $G$, $S$, ${P}_{c}$, ${P}_{m}$, ${fit}^{\star}=0$;

        \FOR{$g$ $\leftarrow$ 0 to $G$}
            \STATE Encoding individual genotype ${\theta}_{g}$;
        \ENDFOR

        \FOR{$s$ $\leftarrow$ 0 to $S$}
            \FOR{$g$ $\leftarrow$ 0 to $G$}
                \STATE ${X}_{adv}^{g}=S(X;{\theta}_{g},\mathcal{M})$;
                \STATE ${fit}_{g} = 1-({y}_{obj}^{g} \leftarrow f({X}_{adv}^{g})$;
                \IF{${fit}_{g} > {fit}^{\star}$}
                    \STATE ${\theta}^{\star}\leftarrow{\theta}_{g}$;
                    \STATE ${fit}^{\star}\leftarrow{fit}_{g}$;
                \ENDIF
            \ENDFOR
            \STATE Update: ${POP}_{s+1}\xleftarrow{SL} {fit}_{g}$;
            \STATE Update: ${POP}_{s+1}\xleftarrow{CS} {P}_{c}$;
            \STATE Update: ${POP}_{s+1}\xleftarrow{MT} {P}_{m}$;

        \ENDFOR
        \STATE \textbf{Output:} ${\theta}^{\star}$;

	\end{algorithmic}  
\end{algorithm}

	

            
        


\section{Experiments}
\label{sec4}

\subsection{Experimental setting}

\textbf{Dataset:} In line with AdvIB \cite{AdvIB}, we use the FLIR dataset \cite{FLIR} for training infrared pedestrian detectors and testing digital attacks. The FLIR dataset comprises a total of 10,228 infrared images, captured using a FLIR Tau2 thermal camera. A notable feature of this dataset is the meticulous manual annotation of all thermal images, accurately distinguishing four categories of target objects: people, bicycles, cars, and dogs. For model training, we implement a strict filtering process, retaining only pedestrian images with heights exceeding 120 pixels. This refinement yielded 1,011 highly relevant samples for training infrared pedestrian detectors. Subsequent adversarial attack experiments are conducted using the test set of this dataset, providing a basis for comprehensive digital attack simulations.

\textbf{Object Detectors:} For the object detection task, this study adopts a series of advanced detection models previously utilized by AdvIB, including Yolo v3 \cite{Yolov3}, DETR \cite{DETR}, Mask R-CNN \cite{Mask}, Faster R-CNN \cite{faster}, Libra R-CNN \cite{Libra}, and RetinaNet \cite{Retina}. These models undergo meticulous training on the specially curated dataset to optimize their target recognition capabilities in complex scenarios. After rigorous training processes, the models demonstrated excellent performance on their respective test sets, with the following average precision scores: Yolo v3 achieves 90.7\%, DETR reaches 91.2\%, Mask R-CNN records 89.5\%, Faster R-CNN achieves 90.8\%, Libra R-CNN obtaines 88.0\%, and RetinaNet leads with a high precision of 93.0\%. These results not only validate the robustness of the selected models but also lay a solid foundation for subsequent adversarial attack research and defense strategy development. In subsequent attack experiments, unless otherwise specified, all attacks are conducted against Yolo v3.

\begin{figure*}[h]
\centering
\includegraphics[width=1\linewidth]{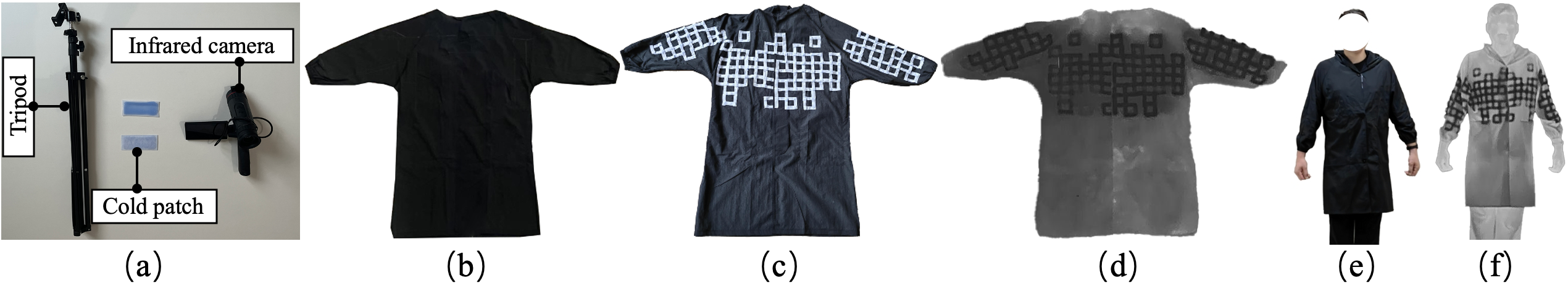} 
\caption{Experimental devices. (a) Tripod, cold patch, and infrared camera. (b) Visible spectrum image of the outer surface of the clothing. (c) Visible spectrum image of the inner surface of the clothing. (d) Infrared image of the clothing. (e) Infrared image of a pedestrian. (f) Visible spectrum image of a pedestrian.}
\vspace{-0.3cm}
\label{figure4}
\end{figure*}

\textbf{Experimental Devices:} During the physical attack experiments, our devices, as shown in Figure \ref{figure4} (a), consist of a stable tripod, a high-performance infrared camera, and cold patches. The infrared camera used is the HIKMICRO FQ25, which boasts excellent imaging performance with the following specifications: a focal plane array (FPA) of 640×512 pixels and a noise equivalent temperature difference (NETD) of less than 25 mK, ensuring high detection sensitivity and image clarity. To verify the general applicability of the method, we test the AdvGrid technique on various infrared camera models. The results clearly indicated that the effectiveness of the technique is not dependent on the specific camera model, demonstrating good compatibility and versatility. The cold patch, capable of maintaining a stable temperature at 24°C for up to 10 hours, provides reliable temperature control for long-duration experiments. Figures \ref{figure4} (b), (c), and (d) show the visible light images of the external and internal views of the infrared stealth clothing generated by our method, as well as the infrared image. It is evident that without the aid of a infrared camera, the perturbations are imperceptible to the naked eye. Figures \ref{figure4} (e) and (f) respectively display the visible light and infrared images of a person wearing the stealth clothing.

\textbf{Evaluation Metrics:} The goal of the proposed AdvGrid is to achieve evasion attacks. Therefore, the attack success rate is used to evaluate the effectiveness of the proposed method, the higher the attack success rate, the more adversarial the method, and vice versa. The formula for calculating the attack success rate is as follows:

\begin{equation}
\label{eq:Positional Encoding}
\begin{split}
    &{\rm ASR}(X) = 1-\frac{1}{N}\sum_{n=1}^{N}F({y}_{obj}^{n})\\
    &F({y}_{obj}^{n})=
        \begin{cases}
        0 & {y}_{obj}^{n} < 0.5 \\
        1 & otherwise
        \end{cases}
\end{split}
\end{equation}
where $N$ denotes the number of true positive labels in the dataset that the detector can identify in the absence of any attacks. Throughout all our attack experiments, a threshold of 0.5 is maintained. Specifically, an evasion attack is deemed successful if the confidence level of the detected target drops below this threshold.

\textbf{Baseline Methods:} Since AdvGrid is a black-box attack, this method uses existing black-box attacks as baseline methods for comparison, including HCB, AdvIB, and AdvIC.

\textbf{Other Settings:} In this method, the dimension of the infrared grid is set to $D=8$, and the width is set to one-fifth of the height of the infrared pedestrian target detection box. The genetic algorithm parameters are set as follows: $G=50$ (population size), $S=10$ (number of iterations), ${P}_{c}=0.6$ (crossover rate), and ${P}_{m}=0.1$ (mutation rate). All adversarial attack experiments are run on an NVIDIA GeForce RTX 4090 GPU.

\subsection{Evaluation of effectiveness}

\begin{figure*}[h]
\centering
\includegraphics[width=1\linewidth]{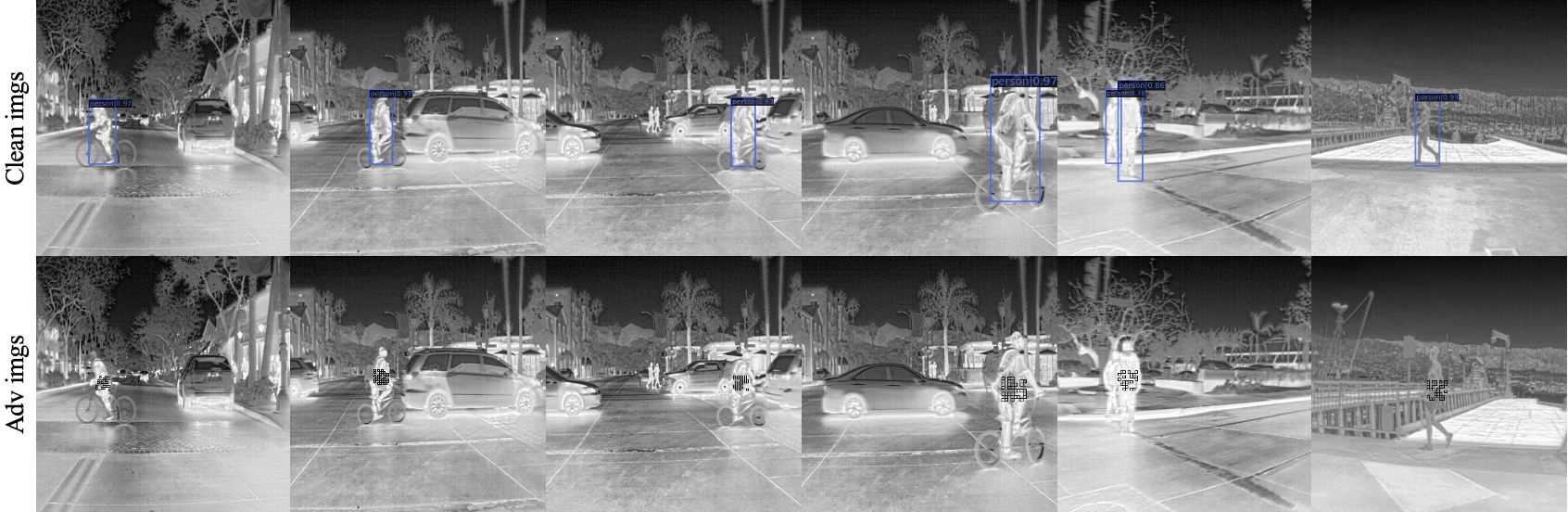} 
\caption{Digital samples generated by AdvGrid.}
\vspace{-0.3cm}
\label{figure5}
\end{figure*}

Digital Testing: To evaluate the effectiveness of our digital attacks, we carefully select specific data from the FLIR dataset to form our test set and implement our attack strategy based on this selection. Using a meticulously designed infrared grid attack configuration (with grid dimensions of 8 and a width equal to one-fifth of the detection box height), we achieve an 80\% attack success rate with an average query count of only 117.53. These experimental results robustly demonstrate the effectiveness and efficiency of our proposed method in a digital environment. Figure \ref{figure5} vividly illustrates the adversarial samples generated during this attack process. As shown, the infrared grid perturbations are seamlessly integrated into the pedestrian images, maintaining a degree of visual stealthiness while effectively evading infrared pedestrian detection systems. This underscores the potential and advantages of our method in practical applications.

\begin{figure*}
\centering
\includegraphics[width=1\linewidth]{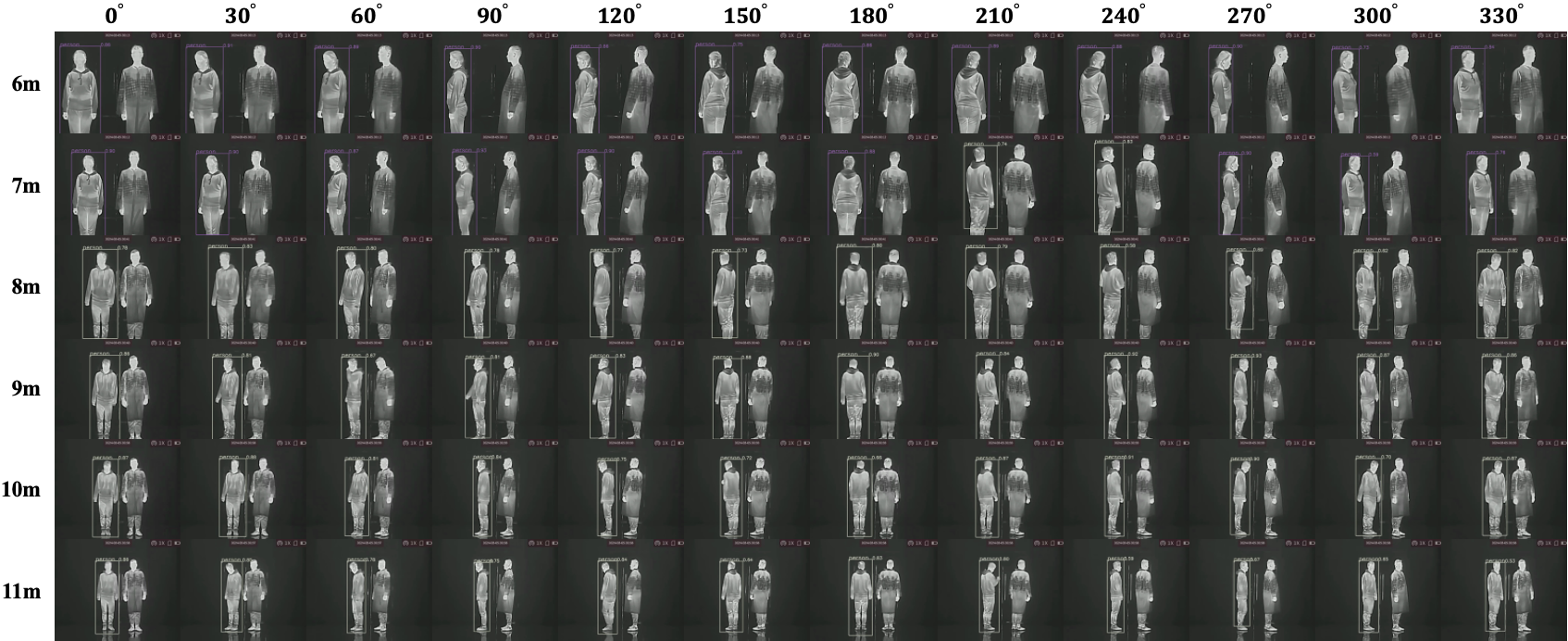} 
\caption{Physical samples generated by AdvGrid.}
\vspace{-0.3cm}
\label{figure6}
\end{figure*}

Physical Testing: In the physical testing phase, we use carefully crafted infrared stealth clothing to conduct the attack tests. The testing process involved pedestrians wearing the stealth clothing and using an infrared camera to capture images from various angles and distances, ranging from 6 meters to 11 meters in 1-meter intervals, and covering a 360-degree field of view with 30-degree intervals at each stop. We meticulously analyze the attack success rates for each combination of distance and angle, compiling all data into Table \ref{tab2}. Key findings from this data include: 1) Overall Excellent Performance: Across all physical tests, we achieve an average attack success rate of 91.86\%, strongly validating the applicability and effectiveness of our proposed method in real-world scenarios. 2) Multi-Angle Effectiveness: Regardless of distance or angle, our method consistently produces significant adversarial effects. Even under the most challenging conditions, we achieve at least a 49.26\% attack success rate, demonstrating robustness in various settings. 3) View Angle Adaptability: Remarkably, nearly half (34 out of 72) of our test cases achieve a perfect 100\% attack success rate, highlighting the method's powerful capability to handle multi-angle challenges and validating its reliability in multi-view adversarial attacks. To visually present these test results, Figure \ref{figure6} showcases a collection of physical samples generated by our method, encompassing adversarial examples from various distances and angles. As illustrated, our method successfully produces effective adversarial samples across all viewpoints, further confirming its excellent performance in physical environments.

\begin{table}
	\centering
 
    \setlength{\belowcaptionskip}{5pt}
    \caption{Experimental results of AdvGrid physical attacks against Yolo v3.}
    \label{tab2}
    \resizebox{\columnwidth}{!}{
	\begin{tabular}{ccccccccccccc}

    \hline
    ~&${0}^{\circ}$&${30}^{\circ}$&${60}^{\circ}$&${90}^{\circ}$&${120}^{\circ}$&${150}^{\circ}$&${180}^{\circ}$&${210}^{\circ}$&${240}^{\circ}$&${270}^{\circ}$&${300}^{\circ}$&${330}^{\circ}$\\
    \hline
    6m&100.0&100.0&69.2&93.3&85.7&100.0&65.7&87.9&95.2&91.5&79.0&100.0\\
    7m&100.0&100.0&84.1&94.3&100.0&100.0&100.0&100.0&100.0&66.1&63.2&97.5\\
    
    8m&100.0&100.0&100.0&93.9&100.0&100.0&97.1&100.0&100.0&97.1&88.5&100.0\\
    
    9m&100.0&86.8&96.9&81.5&90.2&98.7&100.0&100.0&100.0&93.9&93.5&100.0\\
    
    10m&82.5&82.7&100.0&93.4&96.2&100.0&100.0&100.0&100.0&71.5&73.2&94.8\\
    
    11m&81.2&98.5&100.0&100.0&86.5&100.0&100.0&49.3&59.2&87.6&100.0&97.5\\
    \hline

\end{tabular}
}
\end{table}

\subsection{Evaluation of stealthiness}

\begin{figure*}
\centering
\includegraphics[width=1\linewidth]{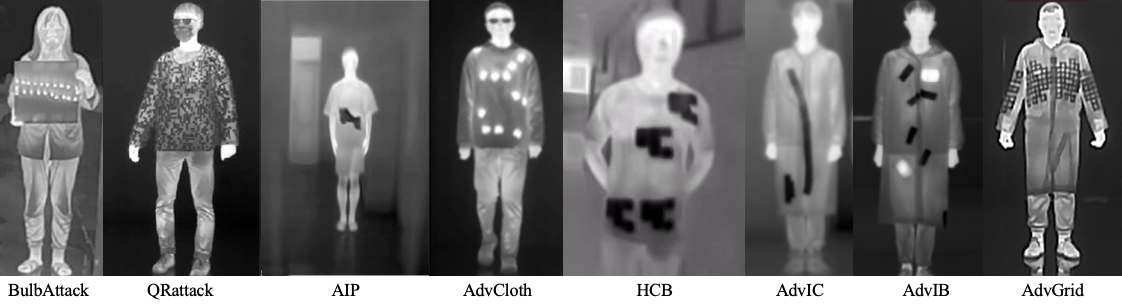} 
\caption{Comparison of physical samples generated by the proposed method and the baseline methods.}
\label{figure7}
\end{figure*}

Observing subfigures (e) and (f) of Figure \ref{figure4}, we notice that, compared to methods such as BulbAttack, QRattack, and AIP, which place perturbations conspicuously on the exterior of the pedestrian's clothing, our method integrates perturbations subtly within the clothing. This design strategy ensures that, without specialized equipment like thermal imaging cameras, ordinary observers cannot easily detect the presence of perturbations, significantly enhancing the stealthiness of adversarial samples in real-world settings. Figure \ref{figure7} further displays the physical adversarial samples generated by baseline methods. Notably, the physical samples generated by AdvGrid exhibit more dispersed perturbation effects, which not only enhance the samples' stealthiness but also give AdvGrid a distinct advantage in physical environments. In stark contrast, the physical samples from baseline methods often appear less covert due to overly concentrated or exposed perturbations.

\subsection{Evaluation of robustness}

\textbf{Deploying AdvGrid to attack various infrared pedestrian target detectors:} We apply AdvGrid to a series of advanced infrared object detectors, including DETR, Mask R-CNN, Faster R-CNN, Libra R-CNN, and RetinaNet. These detectors, after careful tuning, achieve average precision rates of 91.2\%, 89.5\%, 90.8\%, 88.0\%, and 93.0\%, respectively, on a curated infrared dataset, demonstrating their outstanding performance in infrared pedestrian detection tasks. After deploying AdvGrid for the attack, we record the relevant experimental results in Table \ref{tab3} and summarize them as follows: 1) Average attack success rate: AdvGrid achieves an average attack success rate of 59.66\% across all tested detectors. This result not only validates AdvGrid's generalization ability against different infrared pedestrian target detectors but also demonstrates its potential as an effective adversarial strategy. 2) Efficiency analysis: In all conducted attack experiments, AdvGrid requires fewer than 500 queries to complete the attacks. This indicates that AdvGrid not only effectively launches attacks on detectors but also performs quite efficiently in terms of computational resource consumption, making it an ideal choice for deployment in practical scenarios. 3) Attack effects on DETR: It is worth noting that AdvGrid's effectiveness is slightly inferior when attacking DETR detectors based on Transformer architecture compared to other models. This finding suggests that Transformer-based detectors may exhibit stronger robustness and defense capabilities against specific types of adversarial attacks. In summary, AdvGrid demonstrates significant generalization performance and efficiency in adversarial attacks against infrared pedestrian target detectors, especially when targeting non-Transformer architecture detectors. However, for Transformer-based detectors like DETR, its attack effectiveness is slightly lower, indicating directions for future research to improve the universality and effectiveness of adversarial strategies.

\begin{table} [h]
	\centering
 
    \setlength{\belowcaptionskip}{5pt}
    \caption{Experimental results of deploying AdvGrid to attack various advanced detectors.}
    \label{tab3}
	\begin{tabular}{cccc}

    \hline
    $f$&Backbone &ASR &Query\\
    \hline
    Yolo v3&Darknet53 \cite{Yolov3}&80.00&117.53\\
    DETR&Resnet50 \cite{ref1}&6.06&470.23\\
    Mask R-CNN&Resnet50 \cite{ref1}&82.24&100.02\\
    Faster R-CNN&Resnet50 \cite{ref1}&87.16&71.01\\
    Libra R-CNN&Resnet50 \cite{ref1}&75.00&139.55\\
    RetinaNet&Resnet50 \cite{ref1}&27.52&367.78\\
    \hline
    Average &-&59.66&211.02\\
    \hline

\end{tabular}
\end{table}

\textbf{Comparison of AdvGrid with baseline methods' experimental results:} We conduct a comprehensive comparison between the proposed AdvGrid and baseline methods, as shown in Table \ref{tab4}. It can be observed that while the proposed method ranks second to AdvIC in digital attack tests, it outperforms all baseline methods in physical attack tests. Additionally, although AdvGrid's digital attack performance is not as good as AdvIC's, AdvIC lacks multi-view adversarial effects.

\begin{table}[h]
\setlength{\belowcaptionskip}{5pt}
\centering
\caption{Experimental results of the proposed method compared with the baseline methods.}
\label{tab4}
\begin{tabular}{ccccc}

\hline

\multirow{2}*{Method} & \multicolumn{2}{c}{Digital} & Physical&\multirow{2}*{Multi-view}\\
\cmidrule(r){2-3}
\cmidrule(r){4-4}

~ & ASR & Query & ASR & ~\\

\hline

HCB \cite{HCB} & 37.30 & 694.0 & 64.80 & \ding{55}\\

AdvIB  \cite{AdvIB} & 52.30 & 522.40 & 86.34 & \ding{55}\\

AdvIC  \cite{AdvIC} & \textbf{76.20} & \textbf{165.70} & 67.20 & \ding{55}\\

AdvGrid (Ours) & 59.66 & 211.02 & \textbf{91.86} & \ding{51}\\
\hline

\end{tabular}
\end{table}

In conclusion, the proposed method exhibits good generalization performance, and its physical adversarial effects outperform those of baseline methods, confirming the robustness of AdvGrid.

\section{Discussion}

\subsection{Ablation study}

In this section, we conduct ablation experiments on the physical parameters involved in AdvGrid to study the impact of different physical parameters on the adversarial effectiveness of AdvGrid. These experiments include color ablation, dimension ablation, and width ablation, which will be detailed below.

\textbf{Ablation of color:} In infrared physical attacks, both the image and perturbation are grayscale. Therefore, this ablation experiment sets colors to various grayscale values: (0,0,0), (51,51,51), (102,102,102), (153,153,153), (204,204,204), and (255,255,255), to investigate their effects on AdvGrid's adversarial effectiveness. The experimental results summarized in Table \ref{tab5} indicate that AdvGrid achieves its best adversarial effectiveness when the color is (0,0,0) (black), and its effectiveness is weakest when the color is (153,153,153). This can be explained by the fact that when the color is (0,0,0), the perturbation color is most distinct from the color of pedestrians in clean samples, resulting in the best adversarial effectiveness. Conversely, when the color is (153,153,153), the perturbation color is closer to the color of pedestrians in clean samples, resulting in weaker effectiveness.

\begin{table*} 
	\centering
 
    \setlength{\belowcaptionskip}{5pt}
    \caption{Ablation study of color.}
    \label{tab5}
    \resizebox{\columnwidth}{!}{
	\begin{tabular}{ccccccc}

    \hline
    Color&$(0,0,0)$&$(51,51,51)$&$(102,102,102)$&$(153,153,153)$&$(204,204,204)$&$(255,255,255)$\\
    \hline
    ASR&80.00&66.67&47.22&27.78&48.89&56.67\\

    Query&117.53&172.92&275.97&367.91&266.51&228.56\\

    \hline

    \end{tabular}
}
\vspace{0.3cm}
\end{table*}

\textbf{Ablation of dimension:} This experiment explores the impact of the infrared grid dimension on the adversarial effectiveness of AdvGrid. The dimension ranges from 2 to 14, increasing in increments of 2. Smaller dimensions lead to fewer grid cells, reduced perturbation within the grid, and larger blank areas, while larger dimensions have the opposite effect. The experimental results, summarized in Table \ref{tab6}, indicate that AdvGrid's attack success rate initially rises with increasing dimensions, peaking at a dimension of 4, after which it declines with further increases. In practical physical experiments, we opt for a dimension of 8 instead of 4, as a dimension of 4 results in larger blank areas within the grid, hindering effective side-view adversarial attacks due to significant perturbation deformation. Thus, a dimension of 8 is considered the optimal configuration for physical attacks in real-world scenarios.

\begin{table*} [h]
	\centering
 
    \setlength{\belowcaptionskip}{5pt}
    \caption{Ablation study of dimension.}
    \label{tab6}
	\begin{tabular}{cccccccc}

    \hline
    $D$&2&4&6&8&10&12&14\\
    \hline
    ASR&40.56&83.89&82.78&80.00&73.89&71.67&60.00\\

    Query&297.92&93.97&97.01&117.53&135.06&150.92&209.48\\

    \hline

    \end{tabular}
\vspace{0.3cm}
\end{table*}

\textbf{Ablation of width:} This experiment studies the effect of the overall width of the infrared grid on AdvGrid's adversarial effectiveness. The width is set to 1/3, 1/4, 1/5, 1/6, 1/7, and 1/8 of the height of the pedestrian detection box. A wider grid results in more perturbation and poorer stealthiness, and vice versa. The experimental results summarized in Table \ref{tab7} show that AdvGrid's attack success rate increases with the width of the grid. In this experiment, selecting a grid width of 1/5 of the height of the pedestrian detection box is a reasonable choice. When the width continues to increase, the perturbation will exceed the pedestrian target area.

\begin{table*} [h]
	\centering
 
    \setlength{\belowcaptionskip}{5pt}
    \caption{Ablation study of width.}
    \label{tab7}
	\begin{tabular}{ccccccc}

    \hline
    $W$&1/3&1/4&1/5&1/6&1/7&1/8\\
    \hline
    ASR&97.78&88.89&80.00&68.33&59.44&54.88\\

    Query&17.41&67.58&117.53&170.58&212.53&225.71\\

    \hline

    \end{tabular}
\end{table*}

\subsection{Deploying AdvGrid to attack infrared vehicle detectors}

\begin{figure*}
\centering
\includegraphics[width=1\linewidth]{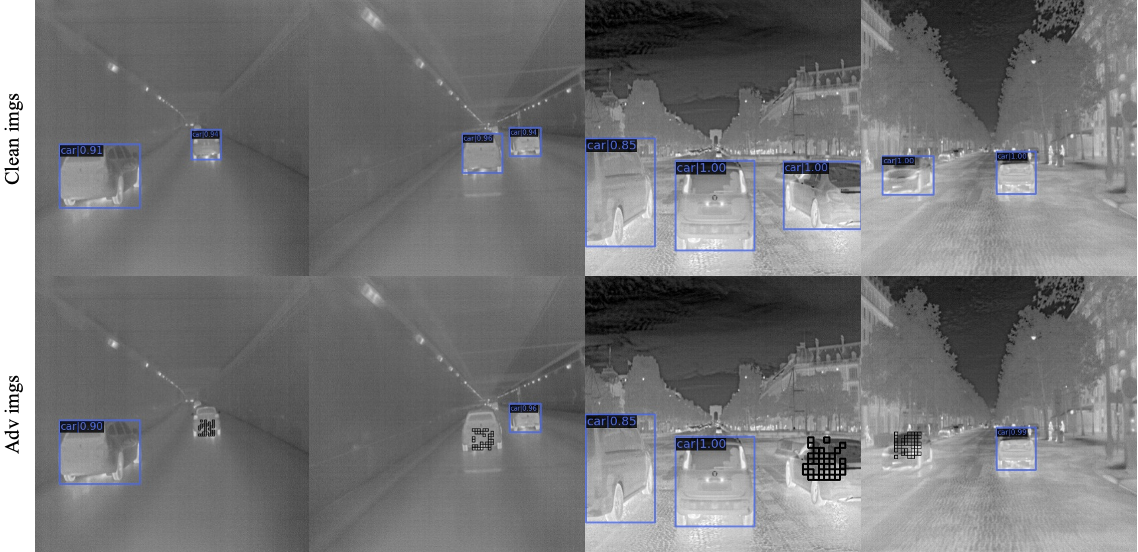} 
\caption{Display of adversarial samples generated by deploying AdvGrid to attack vehicle target detectors.}
\label{figure8}
\end{figure*}

In addition to attacking infrared pedestrian detection systems with AdvGrid, we conduct experiments to validate its applicability for attacking infrared vehicle detection systems, further verifying the generalization performance of the proposed method. We fine-tune infrared vehicle detection detectors, including Yolo v3, DETR, Mask R-CNN, Faster R-CNN, Libra R-CNN, and RetinaNet, on the FLIR dataset, achieving average precision rates of 92.1\%, 94.6\%, 94.2\%, 94.4\%, 95.6\%, and 95.5\%, respectively, on the test set. We summarize the experimental results of attacking these vehicle detection detectors with AdvGrid in Table \ref{tab8}, from which the following conclusions can be drawn: 1) Overall, the experiment achieves an average attack success rate of 50.78\% with an average query count of 260.91, confirming the method's generalization effectiveness across different tasks. 2) Similar to the results in Table \ref{tab3}, AdvGrid exhibits the weakest adversarial effectiveness when attacking DETR, once again confirming the robustness of Transformer-based object detectors. Figure \ref{figure8} displays adversarial samples generated from this experiment, demonstrating that vehicles after perturbation cannot be recognized by the target detectors. 
This experiment provides further evidence of AdvGrid's versatility and effectiveness in attacking various types of object detection systems, contributing to a comprehensive understanding of its adversarial capabilities.

\begin{table} [h]
	\centering
 
    \setlength{\belowcaptionskip}{5pt}
    \caption{Experimental results of deploying AdvGrid to attack various advanced detectors.}
    \label{tab8}
	\begin{tabular}{cccc}

    \hline
    $f$& Backbone &ASR &Query\\
    \hline
    Yolo v3&Darknet53 \cite{Yolov3}&86.45&79.01\\
    DETR&Resnet50 \cite{ref1}&14.71&439.89\\
    Mask R-CNN&Resnet50 \cite{ref1}&26.50&391.26\\
    Faster R-CNN&Resnet50 \cite{ref1}&76.19&135.14\\
    Libra R-CNN&Resnet50 \cite{ref1}&37.50&325.60\\
    RetinaNet&Resnet50 \cite{ref1}&63.31&194.53\\
    \hline
    Average &-&50.78&260.91\\
    \hline

\end{tabular}
\vspace{0.3cm}
\end{table}

\subsection{Optimizing AdvGrid with different optimization algorithms}

In this section, we compare the effects of different adversarial optimization algorithms on AdvGrid by optimizing AdvGrid with these algorithms. The optimization algorithms used here include random optimization, genetic algorithm optimization \cite{GA}, particle swarm optimization \cite{PSO}, and differential evolution algorithm \cite{DE} optimization. We summarize the experimental results in Table \ref{tab9}, from which it can be observed that among all optimization algorithms, random optimization yields the weakest optimization effect, achieving only a 58.56\% attack success rate. On the other hand, genetic algorithm demonstrated the most outstanding optimization effect in this experiment, achieving an 80.00\% attack success rate.

\begin{table*}
	\centering
 
    \setlength{\belowcaptionskip}{5pt}
    \caption{Experimental results of optimizing AdvGrid using different optimization methods.}
    \label{tab9}
	\begin{tabular}{ccccc}

    \hline
    ~&Random&GA&PSO&DE\\
    \hline
    ASR&58.56&80.00&62.54&74.72\\

    Query&253.29&117.53&188.41&143.26\\

    \hline

    \end{tabular}
\end{table*}




\subsection{Attack transferability of AdvGrid}

To evaluate AdvGrid's transfer attack capability, we use adversarial samples generated by AdvGrid that successfully attacked Yolo v3 as a dataset and conduct transfer attacks against DETR, Mask R-CNN, Faster R-CNN, Libra R-CNN, and RetinaNet. These transfer attacks achieve success rates of 27.03\%, 27.03\%, 35.14\%, 27.03\%, and 18.92\%, respectively. These results indicate that AdvGrid's transfer attacks are effective against most detectors in the digital environment, albeit relatively weaker against RetinaNet.

In physical transfer attack experiments, we utilize physical adversarial samples generated by AdvGrid that successfully attacked Yolo v3 as a dataset and conduct transfer attacks against DETR, Mask R-CNN, Faster R-CNN, Libra R-CNN, and RetinaNet. We summarize these experimental results in Tables \ref{tab10}, \ref{tab11}, \ref{tab12}, \ref{tab13}, \ref{tab14}, and draw the following conclusions: 1) Overall, AdvGrid achieves effective physical transfer attacks from Yolo v3 to other advanced infrared pedestrian detection detectors (except for RetinaNet). 2) Yolo v3 to DETR: While 27 out of 72 cases resulted in a 0\% attack success rate, the best attack effectiveness is observed at a distance of 6 meters. 3) Yolo v3 to Mask R-CNN: Achieves a high success rate, with optimal effectiveness observed at a distance of 6 meters, with very few cases of attack failure. 4) Yolo v3 to Faster R-CNN: Optimal attack effectiveness is observed at shorter distances, weakening as the distance increased. 5) Yolo v3 to Libra R-CNN: Overall, exhibits good attack effectiveness, with better performance observed at closer distances. 6)Yolo v3 to RetinaNet: In most cases, effective attacks could not be achieved, with some effectiveness observed only at closer distances, and attacks almost entirely ineffective at longer distances.
In summary, AdvGrid demonstrates its transfer attack capability in both digital and physical environments, especially against DETR, Mask R-CNN, Faster R-CNN, and Libra R-CNN, where effective transfer attacks can be achieved under specific conditions. However, for RetinaNet, its transfer attack effectiveness is unsatisfactory in both digital and physical environments, particularly ineffective at longer distances in the physical environment. These findings highlight AdvGrid's capability for transfer attacks across different detectors while also revealing its limitations against certain detectors (such as RetinaNet) and in physical environments. These insights are crucial for future improvements in enhancing AdvGrid's universality and optimizing its performance in complex environments.

\begin{table}
	\centering
 
    \setlength{\belowcaptionskip}{5pt}
    \caption{Physical transfer attack from Yolo v3 to DETR.}
    \label{tab10}
	\begin{tabular}{ccccccccccccc}

    \hline
    ~&${0}^{\circ}$&${30}^{\circ}$&${60}^{\circ}$&${90}^{\circ}$&${120}^{\circ}$&${150}^{\circ}$&${180}^{\circ}$&${210}^{\circ}$&${240}^{\circ}$&${270}^{\circ}$&${300}^{\circ}$&${330}^{\circ}$\\
    \hline
    6m&73.1&15.9&100.0&100.0&65.3&48.3&68.3&80.0&80.8&100.0&100.0&41.9\\
    7m&31.0&21.9&8.2&11.5&30.8&30.6&73.5&56.9&17.9&28.4&12.1&36.7\\
    
    8m&0.0&2.0&0.0&0.0&12.5&96.8&56.1&45.6&30.4&43.0&9.2&25.5\\
    
    9m&0.0&0.0&0.0&0.0&2.9&22.2&14.8&0.0&3.8&0.0&0.0&0.0\\
    
    10m&0.0&11.1&0.0&0.0&0.0&6.8&12.9&2.6&0.0&25.8&2.5&0.0\\
    
    11m&0.0&0.0&0.0&0.0&10.2&0.0&1.1&0.0&0.0&0.0&0.0&0.0\\
    \hline

\end{tabular}
\end{table}

\begin{table}
	\centering
 
    \setlength{\belowcaptionskip}{5pt}
    \caption{Physical transfer attack from Yolo v3 to Mask R-CNN.}
    \label{tab11}
	\begin{tabular}{ccccccccccccc}

    \hline
    ~&${0}^{\circ}$&${30}^{\circ}$&${60}^{\circ}$&${90}^{\circ}$&${120}^{\circ}$&${150}^{\circ}$&${180}^{\circ}$&${210}^{\circ}$&${240}^{\circ}$&${270}^{\circ}$&${300}^{\circ}$&${330}^{\circ}$\\
    \hline
    6m&57.7&36.2&58.8&60.0&16.3&51.7&83.2&74.4&71.7&100.0&100.0&100.0\\
    7m&41.4&50.0&38.8&41.4&23.1&37.5&58.8&44.8&33.3&25.4&57.6&70.0\\
    
    8m&33.3&30.6&35.7&31.1&50.0&41.9&39.4&36.8&12.5&54.0&56.9&25.5\\
    
    9m&28.2&26.7&9.7&8.5&45.7&61.1&70.4&25.9&61.5&18.4&29.1&20.3\\
    
    10m&24.1&4.4&18.8&17.4&85.6&40.5&57.0&0.9&37.0&58.1&11.4&0.0\\
    
    11m&0.0&3.9&3.5&0.0&56.0&50.0&60.2&26.3&75.0&22.4&0.0&0.0\\
    \hline

\end{tabular}
\end{table}

\begin{table}
	\centering
 
    \setlength{\belowcaptionskip}{5pt}
    \caption{Physical transfer attack from Yolo v3 to Faster R-CNN.}
    \label{tab12}
	\begin{tabular}{ccccccccccccc}

    \hline
    ~&${0}^{\circ}$&${30}^{\circ}$&${60}^{\circ}$&${90}^{\circ}$&${120}^{\circ}$&${150}^{\circ}$&${180}^{\circ}$&${210}^{\circ}$&${240}^{\circ}$&${270}^{\circ}$&${300}^{\circ}$&${330}^{\circ}$\\
    \hline
    6m&76.1&29.0&98.5&100.0&40.1&44.8&36.6&51.1&45.5&100.0&100.0&100.0\\
    7m&37.9&40.6&36.7&57.5&53.8&48.6&38.2&37.9&29.5&22.4&9.1&26.7\\
    
    8m&37.3&36.7&17.9&0.0&16.1&58.1&13.6&26.3&30.4&3.0&20.0&33.3\\
    
    9m&0.0&2.2&4.8&4.9&22.9&30.6&55.6&37.0&15.4&1.0&6.8&20.3\\
    
    10m&38.3&10.0&2.1&0.0&4.8&52.0&29.0&3.5&14.1&33.9&44.3&23.6\\
    
    11m&78.7&35.2&49.1&0.0&0.0&39.4&74.2&78.9&94.6&5.6&2.0&3.5\\
    \hline

\end{tabular}
\end{table}

\begin{table}
	\centering
 
    \setlength{\belowcaptionskip}{5pt}
    \caption{Physical transfer attack from Yolo v3 to Libra R-CNN.}
    \label{tab13}
	\begin{tabular}{ccccccccccccc}

    \hline
    ~&${0}^{\circ}$&${30}^{\circ}$&${60}^{\circ}$&${90}^{\circ}$&${120}^{\circ}$&${150}^{\circ}$&${180}^{\circ}$&${210}^{\circ}$&${240}^{\circ}$&${270}^{\circ}$&${300}^{\circ}$&${330}^{\circ}$\\
    \hline
    6m&60.9&66.7&100.0&100.0&51.7&3.5&1.0&56.7&48.5&100.0&60.0&61.3\\
    7m&13.1&3.1&10.2&67.8&20.5&13.9&26.5&36.2&47.4&50.7&42.4&23.3\\
    
    8m&21.6&4.1&41.1&67.2&16.1&29.0&48.5&49.1&85.7&88.0&27.7&27.5\\
    
    9m&7.7&2.2&14.5&48.8&62.9&59.7&18.5&70.4&87.2&48.0&12.6&0.0\\
    
    10m&12.1&4.4&39.6&50.0&84.8&12.8&47.3&4.4&40.2&86.1&74.7&16.9\\
    
    11m&20.8&9.4&17.5&13.3&8.2&6.4&53.8&0.0&82.4&47.1&6.5&27.1\\
    \hline

\end{tabular}
\end{table}

\begin{table}
	\centering
 
    \setlength{\belowcaptionskip}{5pt}
    \caption{Physical transfer attack from Yolo v3 to RetinaNet.}
    \label{tab14}
	\begin{tabular}{ccccccccccccc}

    \hline
    ~&${0}^{\circ}$&${30}^{\circ}$&${60}^{\circ}$&${90}^{\circ}$&${120}^{\circ}$&${150}^{\circ}$&${180}^{\circ}$&${210}^{\circ}$&${240}^{\circ}$&${270}^{\circ}$&${300}^{\circ}$&${330}^{\circ}$\\
    \hline
    6m&36.4&26.1&4.4&0.0&66.7&62.1&56.4&48.9&45.5&17.4&0.0&0.0\\
    7m&0.0&0.0&0.0&0.0&0.0&5.6&0.0&0.0&0.0&0.0&0.0&0.0\\
    
    8m&0.0&0.0&0.0&0.0&0.0&12.9&0.0&0.0&0.0&0.0&0.0&0.0\\
    
    9m&0.0&0.0&0.0&0.0&11.4&0.0&0.0&0.0&0.0&0.0&0.0&0.0\\
    
    10m&0.0&0.0&0.0&0.0&0.0&0.0&0.0&0.0&0.0&0.0&0.0&0.0\\
    
    11m&0.0&0.0&0.0&0.0&0.0&0.0&0.0&0.0&0.0&0.0&0.0&0.0\\
    \hline

\end{tabular}
\end{table}

\subsection{Defense of AdvGrid}

In this section, we explore the adversarial defense of AdvGrid to investigate its effectiveness against common adversarial defenses. We employ two common adversarial defense mechanisms: Adversarial Training (AT) \cite{QRattack} and Digital Watermarking (DW) \cite{DW}. For AT, we introduce adversarial samples generated by AdvGrid and clean samples to the adversarial training dataset at a ratio of 5:1 and train the Yolo v3 model. After adversarial training, Yolo v3 achieves an average precision of 90.00\% on the test set. For DW, it contains two defense settings: blind image repair and non-blind image repair. In blind image repair, the defender cannot know the location information of the perturbation in advance and needs to detect the perturbation's position first, then perform image repair in the perturbed area. In non-blind image repair, the defender can obtain the location information of the adversarial perturbation and directly repair the image. In this experiment, we adopt non-blind image repair for adversarial defense.

\begin{table*} [h]
	\centering
 
    \setlength{\belowcaptionskip}{5pt}
    \caption{Evaluation of AdvGrid's adversarial effect under adversarial defense.}
    \label{tab15}
	\begin{tabular}{cccc}

    \hline
    ~&No defense&AT&DW\\
    \hline
    ASR&80.00&22.40&31.58\\

    Query&117.53&393.17&-\\
    
    \hline

    \end{tabular}
\vspace{0.3cm}
\end{table*}

We summarize the experimental results of adversarial defense in Table \ref{tab15}. It can be observed that both AT and DW effectively defended against AdvGrid, reducing its attack success rate by 57.60\% and 48.42\%, respectively. The experimental results also indicate that under defense mechanisms, the adversarial effectiveness of AdvGrid will be weakened. However, despite these excellent defense mechanisms being able to effectively defend against AdvGrid, they cannot provide complete defense.

\section{Conclusion}

In this study, we introduce AdvGrid, a multi-view black-box physical attack tailored for infrared pedestrian detectors, addressing the research gap in multi-view black-box attacks in the infrared modality. Our method employs a grid deployed repeatedly in the infrared spectrum, filling the inner side of pedestrian clothing to achieve multi-view adversarial attacks. For adversarial optimization, we utilize genetic algorithms to optimize the physical parameters of the infrared grid to obtain the most adversarial grid. For robust optimization, we employ EOT 
and TPS 
to robustly enhance perturbations, making them more adaptable to the transition from digital to physical environments. We conduct numerous experiments to validate the effectiveness, stealthiness, and robustness of AdvGrid. The method achieved attack success rates of 80.00\% and 91.86\% in digital and physical environments, respectively, validating its effectiveness. In terms of stealthiness, the method's stealthiness is demonstrated by comparing the physical adversarial samples generated by the baseline methods with those generated by AdvGrid. Regarding robustness, by deploying AdvGrid to attack various state-of-the-art object detectors, we achieve an average attack success rate of no less than 50\%, with the method outperforming baseline methods in physical attacks, validating its robustness.

In the future, we will continue to explore the security issues of infrared computer vision systems and the security of computer vision systems in cross-modal scenarios. Infrared object detectors play crucial roles in security screening, autonomous driving, and other fields. Given the outstanding performance of the proposed method in experimental tests, we call for widespread attention to the proposed method and hope that comprehensive defense strategies can be developed against such attacks.





 \bibliographystyle{elsarticle-harv} 
 \bibliography{A}





\end{document}